\documentclass{article}

\usepackage{PRIMEarxiv}

\usepackage[utf8]{inputenc} 
\usepackage[T1]{fontenc}    
\usepackage{hyperref}       
\usepackage{url}            
\usepackage{booktabs}       
\usepackage{amsfonts}       
\usepackage{nicefrac}       
\usepackage{microtype}      
\usepackage{lipsum}

\usepackage{amsmath}
\usepackage{fancyhdr}       
\usepackage{graphicx}       
\graphicspath{{media/}}     

\title{Comparing Deep Neural Network for Multi-Label ECG Diagnosis From Scanned ECG
}

\author{
  Cuong V. Nguyen$^*$ \\
  College of Engineering and Computer Science \\
  VinUniversity, Hanoi, Vietnam \\
  \texttt{cuong.nv@vinuni.edu.vn} \\
  \And
    Hieu X. Nguyen$^*$ \\
  College of Engineering and Computer Science \\
  VinUniversity, Hanoi, Vietnam\\
  \texttt{21hieu.nx@vinuni.edu.vn} \\
  \AND
  Dung D. Pham Minh \\
  College of Engineering and Computer Science \\
  VinUniversity, Hanoi, Vietnam \\
  \texttt{24dung.dpm@vinuni.edu.vn} \\
    \And
  Cuong D. Do \\
  College of Engineering and Computer Science \\
  VinUni-Illinois Smart Health Center\\
  VinUniversity, Hanoi, Vietnam \\
  \texttt{cuong.dd@vinuni.edu.vn} \\
}

\begin{document}
\maketitle

\def\thefootnote{*}\footnotetext{These authors contributed equally to this work.}
\begin{abstract}
Automated ECG diagnosis has seen significant advancements with deep learning techniques, but real-world applications still face challenges when dealing with scanned paper ECGs. In this study, we explore multi-label classification of ECGs extracted from scanned images, moving beyond traditional binary classification (normal/abnormal). We evaluate the performance of multiple deep neural network architectures, including AlexNet, VGG, ResNet, and Vision Transformer, on scanned ECG datasets. Our comparative analysis examines model accuracy, robustness to image artifacts, and generalizability across different ECG conditions. Additionally, we investigate whether ECG signals extracted from scanned images retain sufficient diagnostic information for reliable automated classification. The findings highlight the strengths and limitations of each architecture, providing insights into the feasibility of image-based ECG diagnosis and its potential integration into clinical workflows.
\end{abstract}


\section{Introduction}

Electrocardiograms (ECGs) play a vital role in diagnosing cardiovascular diseases (CVDs), which remain one of the leading causes of mortality worldwide. The accurate interpretation of ECG signals is crucial for early detection and timely medical intervention. Recent advancements in deep learning have significantly improved ECG-based diagnosis, with models achieving cardiologist-level performance \cite{liu2021deep,nguyen2024transfer,hannun2019cardiologist,ribeiro2020automatic,nguyen2024melep,ptbxl_analysis,10521729}. However, most of these approaches rely on high-quality digital ECG data, limiting their real-world applicability in clinical environments where scanned paper ECGs are still prevalent.

Paper-based ECGs remain widely used due to historical adoption, cost-effectiveness, and compatibility with legacy healthcare systems. However, relying on scanned ECGs introduces new challenges, as they contain image-based artifacts such as noise, distortions, and variations in paper quality, which can affect automated diagnostic accuracy. Traditional binary classification (normal vs. abnormal) methods may not fully capture the complexity of cardiac conditions present in real-world ECGs. Therefore, multi-label classification, where multiple cardiac abnormalities are identified simultaneously, presents a more clinically relevant and challenging problem.

Recent efforts in ECG analysis have explored deep neural networks, including AlexNet, VGG, ResNet, and Vision Transformers, for automated classification of ECGs. However, most of these studies focus on clean, digitally recorded ECGs rather than scanned paper ECGs, leaving a gap in understanding how well these models generalize to real-world clinical settings.

In this study, we conduct a comparative analysis of deep learning models applied to multi-label ECG classification from scanned images, evaluating their effectiveness in identifying multiple cardiac conditions despite image-based distortions. Our goal is to determine whether deep neural networks can **retain diagnostic accuracy when applied to scanned ECGs, which is critical for real-world deployment in automated and telemedicine applications.

\textbf{Key contributions:}
\begin{itemize}
    \item We evaluate multiple deep learning architectures (AlexNet, VGG, ResNet, Vision Transformers) for multi-label ECG classification using scanned ECGs, moving beyond binary classification.
    \item We analyze the impact of image-based artifacts on classification accuracy and assess whether deep neural networks can generalize from digital to scanned ECG data.
    \item We highlight the challenges and feasibility of using scanned ECGs for automated diagnosis, providing insights into their potential integration into clinical workflows.
\end{itemize}

\section{Dataset}




In this work, we printed out the PTB dataset on paper ECG for image classification. The original PTB dataset \cite{ptb-dataset} was collected using a non-commercial PTB prototype recorder with 16 input channels (14 ECG, 1 respiration, 1 line voltage), an input voltage range of ±16 mV (offset compensation up to ±300 mV), 100 Ohm input resistance, and 16-bit resolution (0.5 uV/LSB). Signals were synchronously sampled at 1000 Hz across a 0–1 kHz bandwidth, with noise levels reaching a maximum of 10 uV (pp) or 3 uV (RMS). The dataset comprises 549 records from 290 subjects (ages 17–87, mean 57.2), each represented by one to five recordings. Each record includes 15 simultaneous ECG signals: the standard 12-lead ECG (I, II, III, aVR, aVL, aVF, V1–V6) and the three Frank leads (Vx, Vy, Vz), digitized with 16-bit resolution over a ±16.384 mV range. Higher sampling rates (up to 10 kHz) may be available upon request. Most records include a \texttt{.hea} file containing clinical summaries with demographic information, diagnoses, and, where applicable, medical history, prescribed medications, coronary artery pathology, ventriculography, echocardiographic findings, and hemodynamic data.

\begin{figure}
    \centering
    \includegraphics[width=0.7\linewidth]{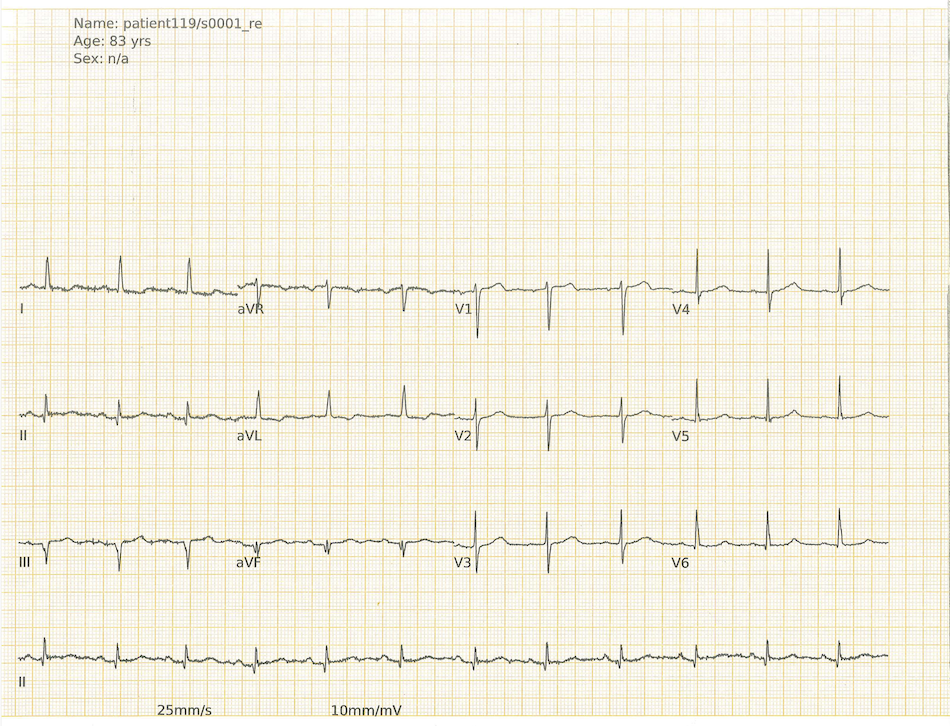}
    \caption{Sample scanned paper ECG}
    \label{fig:sample-scan}
\end{figure}

In this work, we printed 12 out of the 15 leads (I, II, III, aVR, aVL, aVF, V1–V6) from each record in the original PTB dataset onto ECG paper using the ECG Image Kit \cite{ecg-img-kit}. Given the length of the original ECG signals, only a portion of each signal was printed. Consequently, the original digital records (\texttt{.dat} format) were modified to align with the printed segments, ensuring a valid basis for comparison. After printing, all 549 paper ECGs were scanned to complete the dataset collection. Figure \ref{fig:sample-scan} illustrates an example of a scanned ECG, with patient data anonymized.

\section{Indirect Diagnosis via Digitization}

\subsection{Task Description}

The digitization task aims to convert paper-scanned ECGs into digital time-series signals, enabling automated analysis and integration with electronic health records. Standard paper ECGs typically display all 12 leads in 2.5-second segments, arranged in three rows of four leads each. Additionally, an extended 10-second strip of lead II, V1, V2, or V5 is often included at the bottom (see Figure \ref{fig:sample-scan}). The objective is to accurately extract and reconstruct the 12-lead ECG signals from the scanned images, preserving signal morphology and ensuring compatibility with digital processing techniques. The final output consists of digitally reconstructed 12-lead ECG signals, which can be further utilized for clinical evaluation and algorithmic interpretation.

\subsection{Methodology}

In this work, we use VinDigitizer \cite{nguyenvindigitizer}, our approach in the Physionet Challenge 2024, for baseline evaluation. VinDigitizer follows a three-stage pipeline, which consists of detecting signal rows on the paper ECG, extracting the signal from the background, and assigning numerical values to each signal point. 

The first stage focuses on isolating the sections of the image that contain ECG lead signals. Standard paper ECGs typically display all 12 leads in 2.5-second segments, arranged in three rows of four leads each. To facilitate this segmentation, a YOLOv8 \cite{yolov8} model was trained on a dataset of 300 annotated paper ECGs, enabling it to accurately detect and localize the four signal-containing rows.

\begin{figure*}[!ht]
  \centering
  \includegraphics[width=0.5\linewidth]{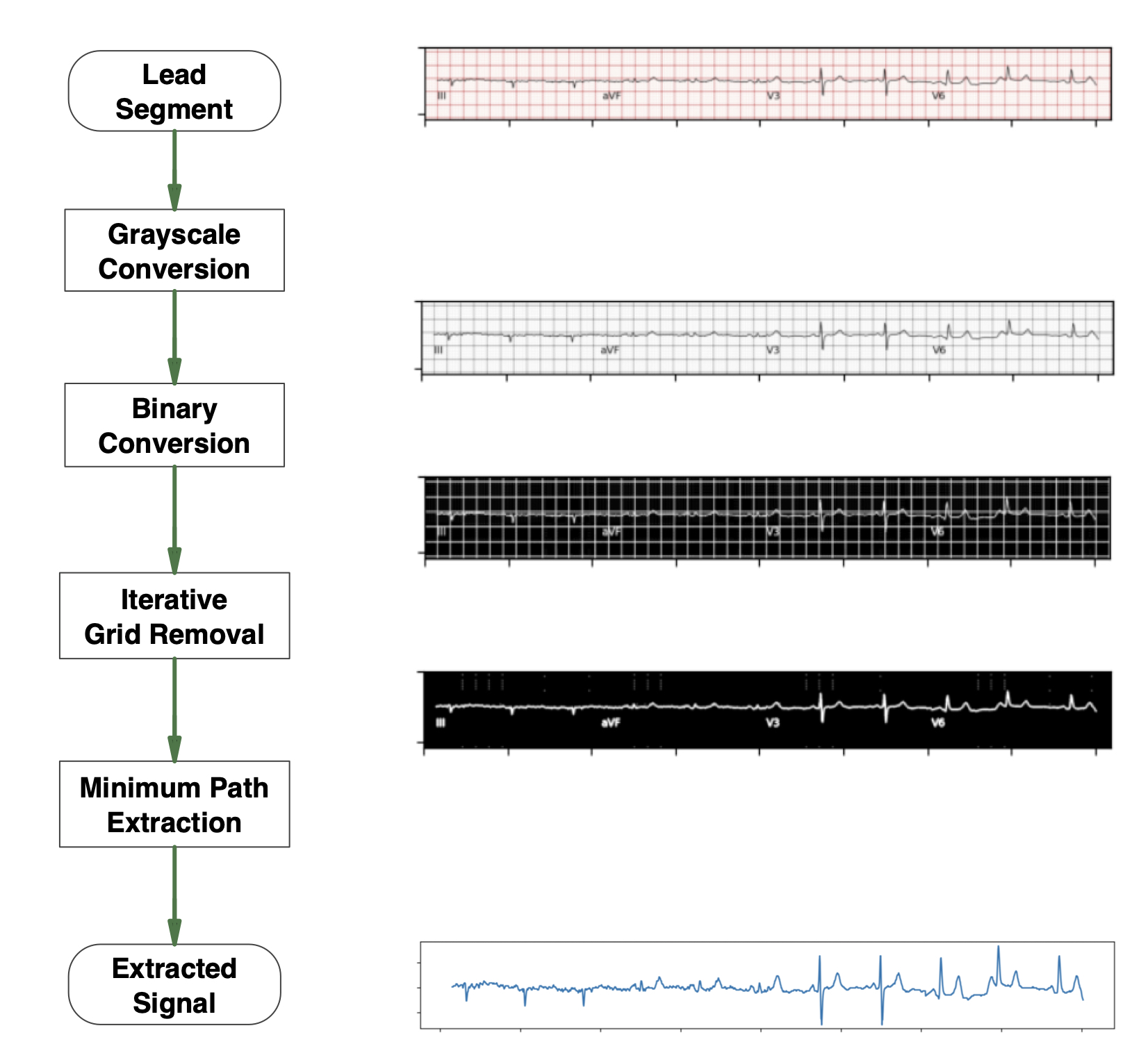}
  \caption{Signal extraction procedure.}
  \label{fig:sig_extraction}
\end{figure*}

The second stage involves extracting numerical representations of the waveform within each segmented ECG row. The signal extraction process, illustrated in Figure \ref{fig:sig_extraction}, begins with converting the RGB image to grayscale, allowing for pixel intensity quantification. The grayscale image is then binarized using Otsu’s thresholding \cite{otsu1975threshold}, where pixels with intensity values below the computed threshold are assigned a value of 0 (black), while those above the threshold are set to 255 (white).

Following binarization, the background grid is iteratively removed using the method outlined in \cite{fortune2022digitizing}. The vertical and horizontal grid lines are detected based on column-wise white-point density. If the grid remains visible in the binary image, the grayscale-to-binary conversion threshold is reduced by 5\%, and the process is repeated until the grid is no longer detectable, resulting in a clean, signal-only binary image.

In the final stage, numerical signal values are extracted from the processed binary image. This step follows the approach proposed in \cite{fortune2022digitizing}, where vertical white pixel clusters are treated as candidate signal nodes, and the waveform is reconstructed by identifying the least-cost path from left to right across the image. The algorithm first detects candidate nodes in each column and then evaluates possible connections to preceding non-empty columns. The cost of each connection is computed based on Euclidean distance and angular alignment, ensuring smooth waveform reconstruction. The path with the lowest total cost is selected as the extracted signal, and an array of vertical pixel coordinates along this path is returned as the final digital representation.

\subsection{Evaluation Metrics}
For the digitization task, we utilized the following metrics: Signal-to-Noise Ratio (SNR), SNR median, Kolmogorov-Smirnov (KS) metric, weighted absolute difference (WAD), and adaptive signed correlation index (ASCI).

Specifically, SNR (in dB) is defined as the ratio of the power of the original signal to the power of the noise, which is the difference between the original and the digitized signal:

\begin{equation}
\text{SNR} = 10 \log_{10} \left( \frac{P_{\text{original}}}{P_{\text{original} - \text{digitized}}} \right)
\end{equation}

A higher SNR indicates that the digitized signal is closer to the original. If the digitized signal is identical to the original, the SNR is infinity.

\subsection{Results}

\begin{figure*}[!ht]
  \centering
  \includegraphics[width=\linewidth]{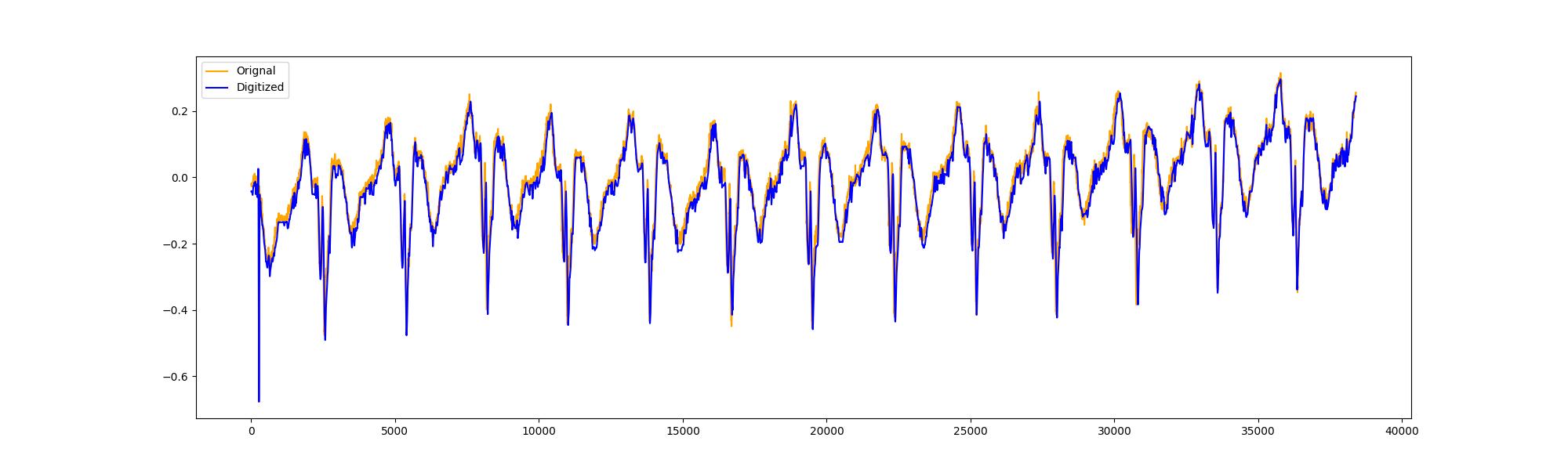}
  \caption{Digitization result of Lead II, patient 001, ID s0010\_re.}
  \label{fig:digitization_result}
\end{figure*}

VinDigitizer was applied to all 549 scanned ECG files from the PTB-Image dataset to convert the printed waveforms back into digital signals. To assess the accuracy of the digitization process, we compared the extracted signals with the corresponding original PTB samples, which had been modified to match the specific portions of the signals that were printed on paper. This ensured a fair and precise evaluation, as the reference signals were aligned with the available printed segments. The signal reconstruction quality was quantified using the signal-to-noise ratio (SNR), yielding a mean SNR of 0.01 dB across the dataset. This result highlights the challenges associated with extracting clean signals from scanned paper ECGs, including variations in printing quality, scanner-induced distortions, and grid removal artifacts. Despite these challenges, the digitized signals maintained recognizable morphological features, demonstrating the feasibility of the proposed method for recovering ECG waveforms from paper records. Figure \ref{fig:digitization_result} presents an example of a digitized ECG signal alongside its original reference.

The digitized ECG can be used for indirect cardiovascular disease diagnosis. There is a direct approach that use the scanned ECG papers for classification, which we explore in the next section.






\section{Direct Diagnosis via Scanned Images}

Work in progress.

\section{Conclusion}

In this study, we investigated the multi-label classification of ECGs from scanned paper images using deep neural networks. Given the prevalence of paper-based ECGs in clinical practice, our research aimed to determine whether deep learning models could accurately diagnose multiple cardiac conditions despite the presence of image artifacts and distortions. We conducted a comparative evaluation of several deep neural network architectures, including AlexNet, VGG, ResNet, and Vision Transformers, analyzing their effectiveness in classifying ECGs extracted from scanned images.

Our results highlight that while deep learning models can achieve promising classification performance on scanned ECGs, their accuracy is affected by the quality of the scanned images and the presence of noise introduced during the printing and scanning process. Among the evaluated models, we observed variations in robustness, with some architectures being more resilient to image-based artifacts than others. These findings suggest that careful preprocessing, model selection, and data augmentation techniques are crucial for improving performance in real-world applications.

Despite the progress made, challenges remain in ensuring the generalizability of deep learning models across different scanning conditions and ECG sources. Future research could explore domain adaptation techniques, self-supervised learning, and hybrid image-signal processing approaches to further enhance classification accuracy. Additionally, incorporating clinically diverse ECG datasets and improving model interpretability could help bridge the gap between automated ECG diagnosis and real-world clinical deployment.

By advancing the use of deep learning for scanned ECG classification, this work contributes to the broader goal of enhancing accessibility to historical ECG data, supporting telemedicine applications, and enabling automated cardiac diagnostics in resource-constrained settings.




\end{document}